\documentclass{article}

\usepackage{PRIMEarxiv}
\usepackage{amsmath} 
\usepackage[utf8]{inputenc} % allow utf-8 input
\usepackage[T1]{fontenc}    % use 8-bit T1 fonts
\usepackage{hyperref}       % hyperlinks
\usepackage{url}            % simple URL typesetting
\usepackage{booktabs}       % professional-quality tables
\usepackage{amsfonts}       % blackboard math symbols
\usepackage{nicefrac}       % compact symbols for 1/2, etc.
\usepackage{microtype}      % microtypography
\usepackage{lipsum}
\usepackage{fancyhdr}       % header
\usepackage{graphicx}       % graphics
\graphicspath{{media/}}     % organize your images and other figures under media/ folder

\usepackage{array}
\newcolumntype{H}{>{\setbox0=\hbox\bgroup}c<{\egroup}@{}}
\usepackage{mathtools}
\usepackage{bbm}
\newcommand{\dataset}{FRED}
  
\title{\dataset: The Florence RGB-Event Drone Dataset}

\author{
  Gabriele Magrini\\
  University of Florence \\
  Florence, Italy\\
  \texttt{gabriele.magrini@unifi.it} \\
   \And
  Niccolò Marini \\
  University of Florence \\
  Florence, Italy\\
  \texttt{niccolo.marini@edu.unifi.it} \\
   \And
  Federico Becattini \\
  University of Siena \\
  Siena, Italy\\
  \texttt{federico.becattini@unisi.it} \\
   \And
   Lorenzo Berlincioni \\
  University of Florence \\
  Florence, Italy\\
  \texttt{lorenzo.berlincioni@unifi.it} \\
   \And
   Niccolò Biondi \\
  University of Florence \\
  Florence, Italy\\
  \texttt{niccolo.biondi@unifi.it} \\
   \And
   Pietro Pala\\
  University of Florence \\
  Florence, Italy\\
  \texttt{pietro.pala@unifi.it} \\
   \And
   Alberto Del Bimbo\\
  University of Florence \\
  Florence, Italy\\
  \texttt{alberto.delbimbo@unifi.it} \\
}

\begin{document}
\maketitle

\begin{abstract}
Small, fast, and lightweight drones present significant challenges for traditional RGB cameras due to their limitations in capturing fast-moving objects, especially under challenging lighting conditions. Event cameras offer an ideal solution, providing high temporal definition and dynamic range, yet existing benchmarks often lack fine temporal resolution or drone-specific motion patterns, hindering progress in these areas.
This paper introduces the Florence RGB-Event Drone dataset (\dataset{}), a novel multimodal dataset specifically designed for drone detection, tracking, and trajectory forecasting, combining RGB video and event streams.
FRED features more than 7 hours of densely annotated drone trajectories, using 5 different drone models and including challenging scenarios such as rain and adverse lighting conditions. We provide detailed evaluation protocols and standard metrics for each task, facilitating reproducible benchmarking. The authors hope FRED will advance research in high-speed drone perception and multimodal spatiotemporal understanding.
\end{abstract}


\maketitle

\section{Introduction}

The ability to accurately detect, track, and forecast the motion of drones is critical for a wide range of applications, including airspace monitoring, collision avoidance, autonomous navigation, and security. These tasks are particularly challenging due to the fast, agile, and often erratic motion of drones, their small size in the visual field, and the need for real-time processing in dynamic environments. Existing benchmarks for object detection and tracking, however, are not geared towards such challenges, often lacking fine temporal resolution or drone-specific motion patterns. %, or support for asynchronous sensing modalities.

While RGB-based drone detection has seen notable progress, it faces several inherent limitations. First, RGB cameras suffer from motion blur during fast movements, which is common in drone flight, degrading detection accuracy. Second, standard frame-based sensing introduces latency and limits temporal resolution, making it difficult to localize fast-moving drones precisely. Third, RGB cameras are sensitive to lighting conditions and can perform poorly in low-light, high-contrast, or high dynamic range environments. These factors hinder robust detection and tracking, motivating the integration of alternative sensing modalities such as event cameras.

Event cameras~\cite{Gallego-2022} offer a promising complement to traditional RGB sensors, providing high temporal resolution, low latency, and high dynamic range by capturing changes in the visual scene as a continuous stream of per-pixel brightness changes. This makes them especially well-suited for tracking fast-moving objects like drones under challenging lighting or motion conditions~\cite{magrini2024neuromorphic, magrini2025flying}.
Despite their growing popularity in robotics and neuromorphic vision, there is currently a lack of large, annotated datasets that combine event and RGB data for the specific purpose of drone perception.

In this paper, we present \dataset{}, a novel multimodal dataset for drone detection, tracking, and trajectory forecasting that jointly leverages RGB video and event streams. The dataset features densely annotated drone trajectories, high-frequency event data, and spatiotemporally aligned RGB frames. The recordings exhibit a high variability, including challenging scenarios like rain, presence of distractors or adverse lighting conditions.
Our design supports flexible supervision, where models may operate with either or both sensing modalities, and accommodates diverse motion patterns and real-world scene complexity.
We provide detailed evaluation protocols and standard metrics for each task, allowing reproducible benchmarking and fair comparison across models.
The dataset is freely available and open source at \url{https://miccunifi.github.io/FRED/}.
We hope this dataset will foster progress in high-speed drone perception and multimodal spatiotemporal understanding.

\begin{figure}[t]
    \includegraphics[width=\textwidth,trim={0 100px 0 100px},clip]{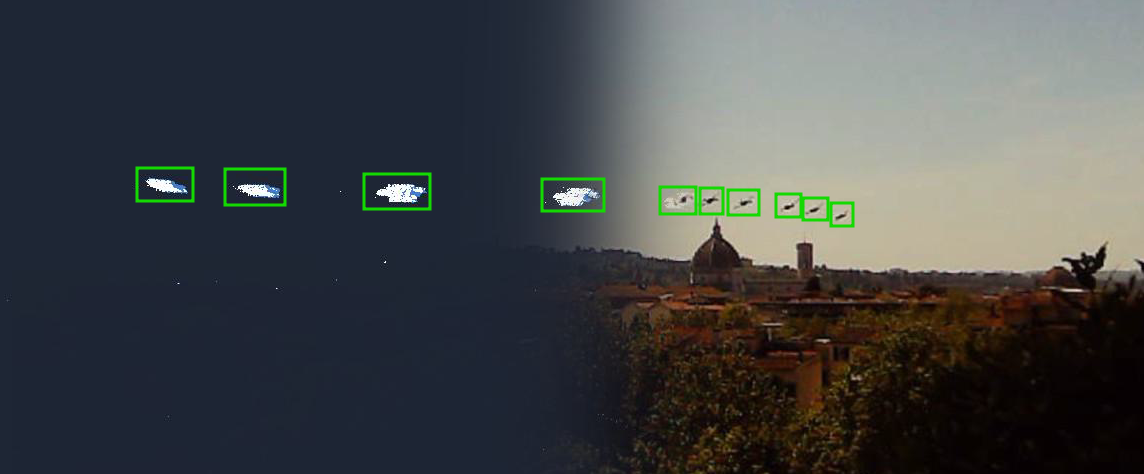}
    \caption{In the Florence RGB-Event Drone dataset (FRED), RGB and Event frames are spatio-temporally synchronized and drones are annotated for detection, tracking and forecasting.}
\label{fig:teaser}
\end{figure}

\section{Related Works}
\paragraph{\textbf{Drone Detection}}
Drone detection has been addressed in several domains, including radio-frequency \cite{al2020drone}, thermal \cite{svanstrom2021real} and acoustic \cite{svanstrom2021real}.
Event cameras, known for their microsecond-level temporal precision and high dynamic range, have gained traction in recent years as a promising tool for drone detection tasks \cite{Stewart-2021, Stewart-2022, Mandula-2024, magrini2024neuromorphic, magrini2025flying}. Unlike general object detection, identifying drones introduces unique challenges. Drones typically have a compact form factor, which makes them difficult to detect, particularly at a distance or within cluttered scenes. Their ability to move rapidly adds another layer of complexity, as it demands real-time responsiveness from detection systems. Furthermore, drones can hover or remain stationary mid-air, causing them to visually merge with the background and reducing motion-based distinguishability.

In prior work \cite{Stewart-2021, Stewart-2022}, a DAVIS sensor was employed to recognize drones by exploiting the event patterns generated by the spinning propellers. The method involves building a frequency histogram from event data, which is then classified between drones and other aerial objects based on high-frequency signatures and subharmonics. However, the effectiveness of this technique heavily depends on the viewpoint since rotating blades produce prominent event patterns only when observed from below. In contrast, classification from oblique or lateral angles remains a significant challenge even for propeller-based systems.
An orthogonal take on the problem, still leveraging a similar point of view, is to establish a virtual fence with neuromorphic cameras to prevent undesired access to restricted areas~\cite{stewart2021drone, lundin2025drone}.

An event cloud-based approach~\cite{magrini2025flying} was recently proposed to discriminate birds from drones with a PointNet~\cite{qi2017pointnet, qi2017pointnet++}. However, the method assumes to have access to an oracle class-agnostic detector.
Other recent approaches~\cite{magrini2024neuromorphic, Mandula-2024} have treated the recognition of drones as an object detection problem.
Mandula et al.~\cite{Mandula-2024} explored a low-power multimodal approach by combining a Prophesee EVK4 event camera and an Raspberry Pi camera on an Nvidia Jetson Xavier NX platform, leveraging a YOLOv5-based model. While this work is among the few to consider a multimodal hardware setup, it lacks detailed insights into the employed fusion strategy and does not report quantitative results on detection accuracy.
Magrini et al.~\cite{magrini2024neuromorphic} deepened the study of multimodal detectors, proposing several DETR-based architectures with different modality fusion strategies, highlighting the effectiveness of the neuromorphic component over its RGB counterpart. Interestingly, jointly leveraging both modalities proves to be beneficial as they capture complementary visual and motion patterns.
Despite this, the model is validated on data with limited variability.

These findings highlight the need for larger-scale multimodal drone datasets. In this paper, we propose a dataset with paired event and RGB data, doubling the size of the largest existing dataset in the literature, including previously unstudied conditions, such as challenging weather conditions and different times of day.
To the best of our knowledge, no existing method has addressed drone tracking and forecasting with an event camera. Our dataset proposes benchmarks for detection, tracking and forecasting.
A comparison between \dataset{} and other existing neuromorphic datasets including drones is shown in Tab.~\ref{table:dataset_comparison}.

\begin{table}[t]
\begin{center}
\resizebox{\columnwidth}{!}{
\begin{tabular}{l|ccccccc}
%\hline
\textbf{Dataset}~ & ~\textbf{Resolution}~ & ~\textbf{RGB/Event}~ & ~\textbf{Duration}~ & ~\textbf{Drone-Centric}& ~\textbf{Rain} & ~\textbf{Day/Night} & ~\textbf{Drone Types}\\
\hline
\textbf{VisEvent \cite{wang2023visevent}} & 346 x 260 & \checkmark/\checkmark & <5h & $\times$ & $\times$  & \checkmark & - \\
\hline
\textbf{EventVOT \cite{wang2024event}}~ &  ~1280 x 720 (HD) & $\times$/\checkmark & <5h & $\times$ &  -  & - & - \\
\hline
\textbf{F-UAV-D \cite{mandula2024towards}} & ~1280 x 720 (HD)& \checkmark/\checkmark & 0h:30m &  \checkmark  & $\times$ & $\times$ & 2 \\
\hline
\textbf{Ev-UAV \cite{LiMiao2024OptoElectronicEngineering}} & ~346 x 240 & $\times$/\checkmark & 0h:15m & \checkmark  & $\times$ & - & - \\
\hline
\textbf{Ev-Flying \cite{magrini2025flying}} & ~1280 x 720 (HD) & $\times$/\checkmark & 1h:07m & $\times$  & $\times$ & $\times$ & 1 \\
\hline
\textbf{NeRDD \cite{magrini2024neuromorphic}} & ~1280 x 720 (HD) & \checkmark/\checkmark & 3h:30m & \checkmark  & $\times$ & $\times$ & 2 \\
\hline
\textbf{\dataset{} (Ours)} & ~1280 x 720 (HD) & \checkmark/\checkmark & 7h:07m & \checkmark & \checkmark  & \checkmark & 5 \\
%\hline
\end{tabular}
}
\caption{Comparison of existing event-based drone datasets. Other datasets either have a low resolution, do not contain RGB versions of the samples, are not Drone-centric or are very small.}
\label{table:dataset_comparison}
\end{center}
\end{table}

\paragraph{\textbf{Event-based Datasets}}
Recent years have seen a notable rise in the release of datasets tailored for event-based object detection \cite{berlincioni2023neuromorphic, Gehrig-2023, Perot-2020}, reflecting the growing interest in neuromorphic vision. Thanks to advances in sensor technology, several high-resolution (Full HD) datasets have emerged \cite{innocenti2021temporal, Perot-2020}. However, datasets that include both event data and time-synchronized RGB frames remain relatively rare. Among the few, some leverage this multimodal setup for improved robustness in object tracking \cite{Wang-2020} or for domain compensation between event and RGB data \cite{Tomy-2022}.

When it comes to drone-specific detection using event data, the landscape becomes even sparser. UAVs are typically included only as a minor category within broader datasets \cite{wang2024event, wang2023visevent}, and dedicated event-based drone datasets are limited. The intersection of three key properties—high resolution, hybrid DVS-RGB streams, and UAV-specific content—is scarcely populated.
%Only few datasets manage to hit this spot.
Mandula et al. \cite{Mandula-2024}  present F-UAV-D, a bimodal dataset of RGB-Event drone recordings with spatio-temporal synchronization, but it comes with significant limitations: it contains only 30 minutes of RGB-event recordings, and the diversity of scenarios is minimal. This lack of variation not only restricts the generalizability of models trained on the data but also impairs their ability to handle complex scenes where multiple dynamic objects may enter and exit the field of view over time.

A more recent work~\cite{magrini2024neuromorphic} presented NeRDD, a more substantial spatio-temporal synchronized Event-RGB dataset containing drone recordings in various scenarios. Nonetheless, the recorded drones are only two medium-sized commercial drones, with limited maneuvering and speed capabilities. The recorded scenarios also do not include challenging scenarios (e.g., rain, nighttime, presence of other flying objects) and most of the videos comprise a single drone.
Whereas the presence of other flying objects has been studied in a follow-up work~\cite{magrini2025flying}, the lack of diversity and larger-scale data availability remains an open issue.
In addition, no tracking or forecasting scenarios are considered in both \cite{magrini2024neuromorphic} and \cite{Mandula-2024}.

In this work, we propose a novel neuromorphic-RGB drone dataset that comprises more than 7 hours of high-resolution, spatio-temporally aligned Event-RGB recordings of 5 different drone models, recorded in semantically complex scenarios and in shifting domains.
%that also integrates the NeRDD dataset presented in \cite{magrini2024neuromorphic}.
Every frame is annotated with drone bounding boxes and track ID, expanding the possible tasks not only to detection but also forecasting and tracking.

\newcommand{\figwidthnew}{.24\textwidth}
\begin{figure*}[t]
    \centering
    \makebox[\textwidth]{\hfill $\longrightarrow$ \textbf{Day to Night} $\longrightarrow$ \hfill}

    \vspace{0.5em}
    \setlength{\tabcolsep}{1pt}
    \begin{tabular}{ccccc}
    \includegraphics[width=\figwidthnew,trim={150px 150px 150px 0},clip]{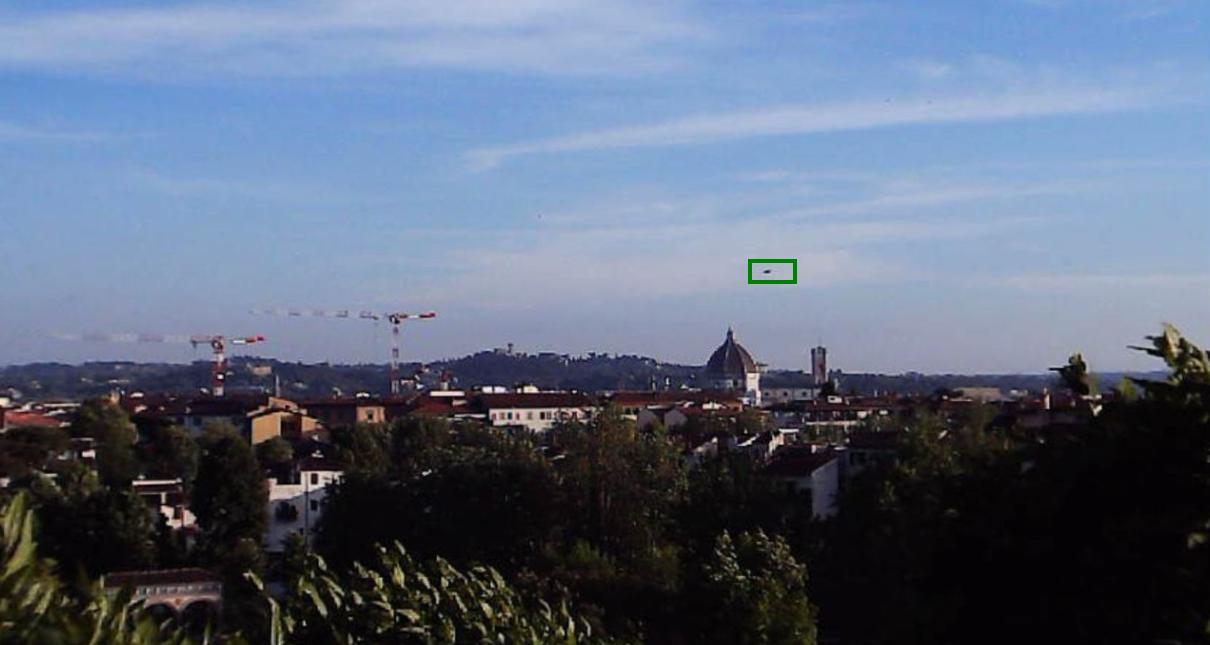}&
    \includegraphics[width=\figwidthnew,trim={150px 150px 150px 0},clip]{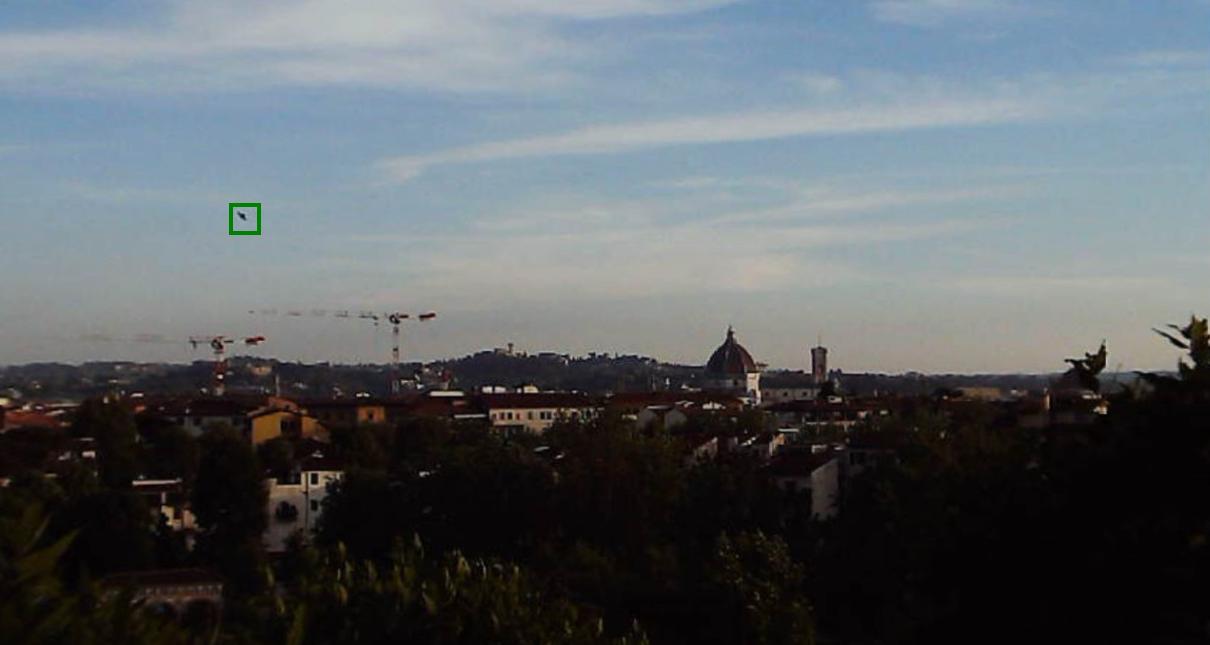} &
   \includegraphics[width=\figwidthnew,trim={150px 150px 150px 0},clip]{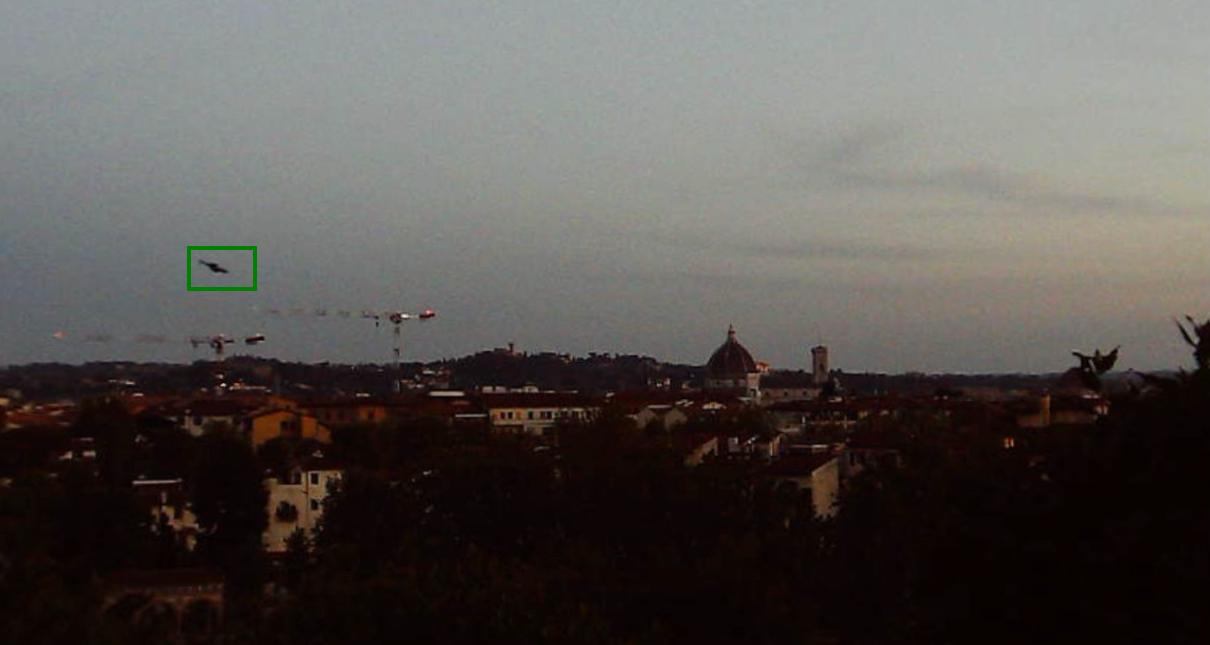} &
    \includegraphics[width=\figwidthnew,trim={150px 150px 150px 0},clip]{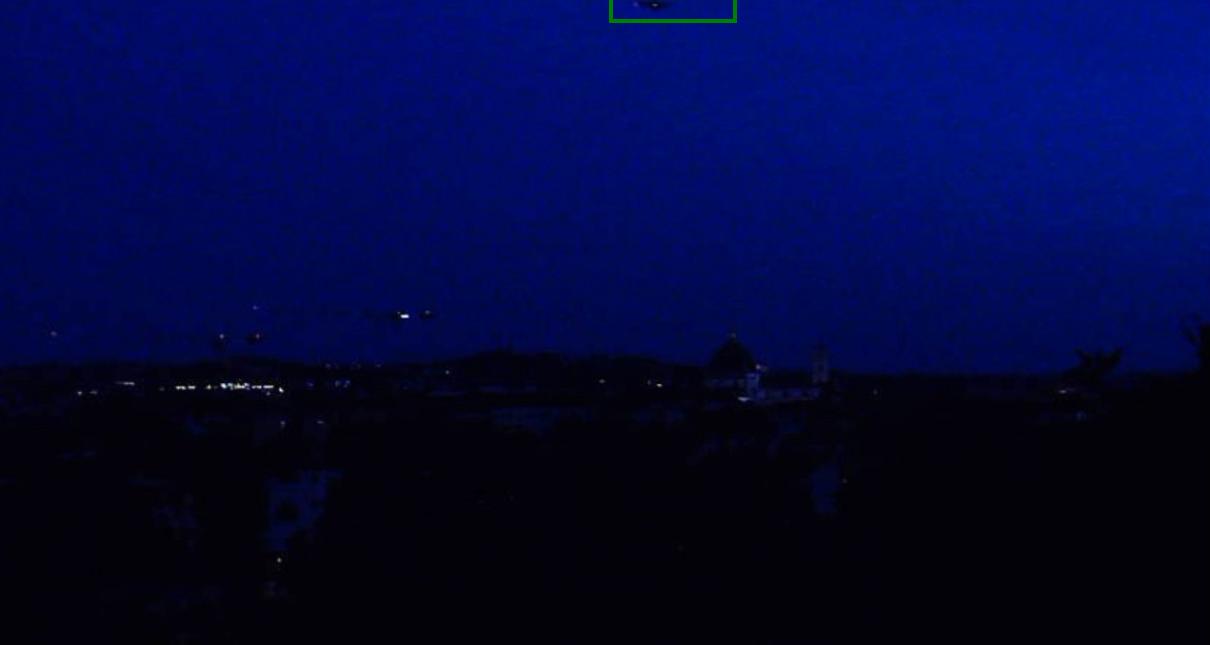} \\
    
    \includegraphics[width=\figwidthnew,trim={150px 150px 150px 0},clip]{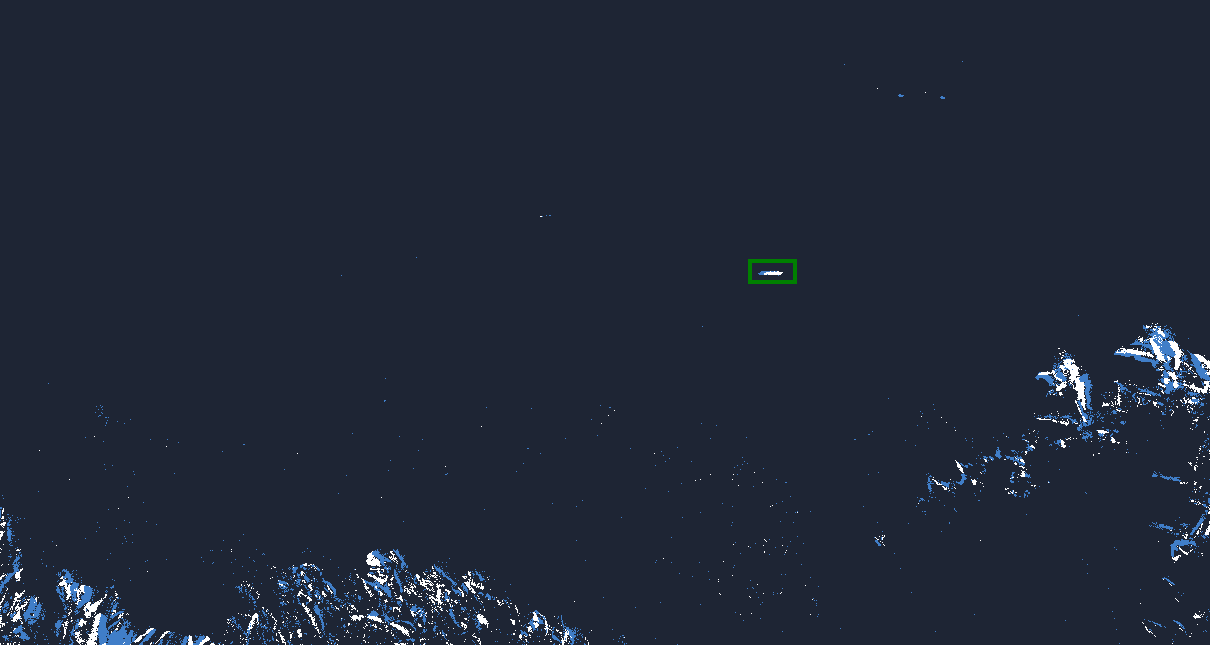} &
    \includegraphics[width=\figwidthnew,trim={150px 150px 150px 0},clip]{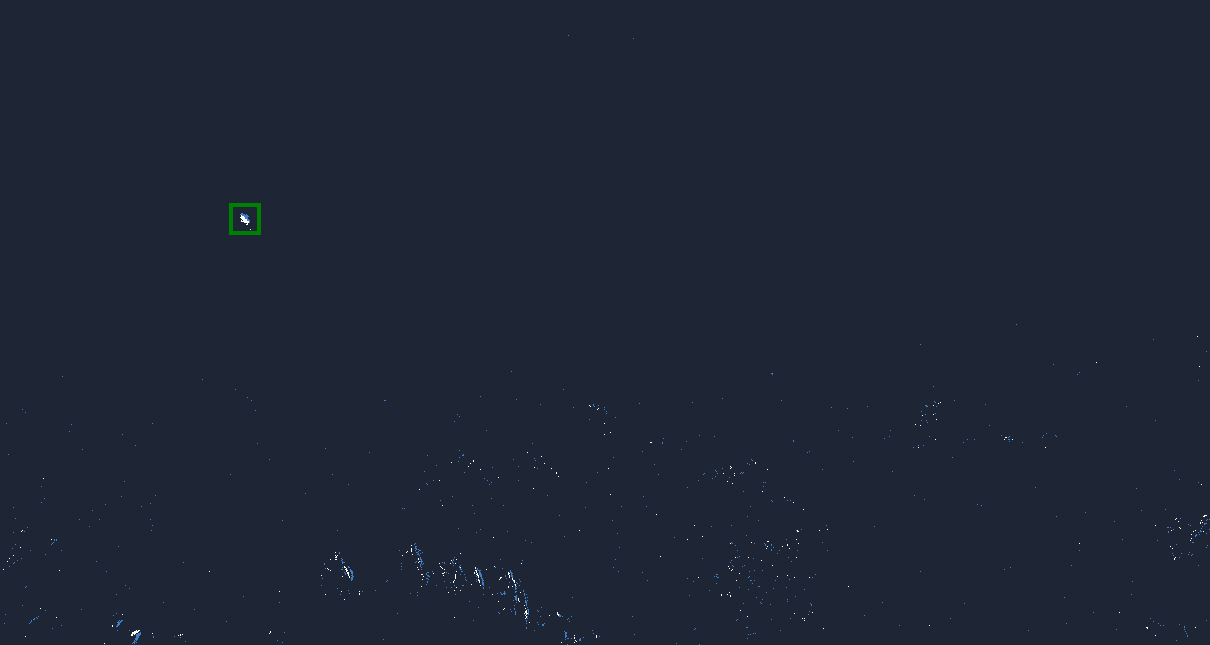} &
    \includegraphics[width=\figwidthnew,trim={150px 150px 150px 0},clip]{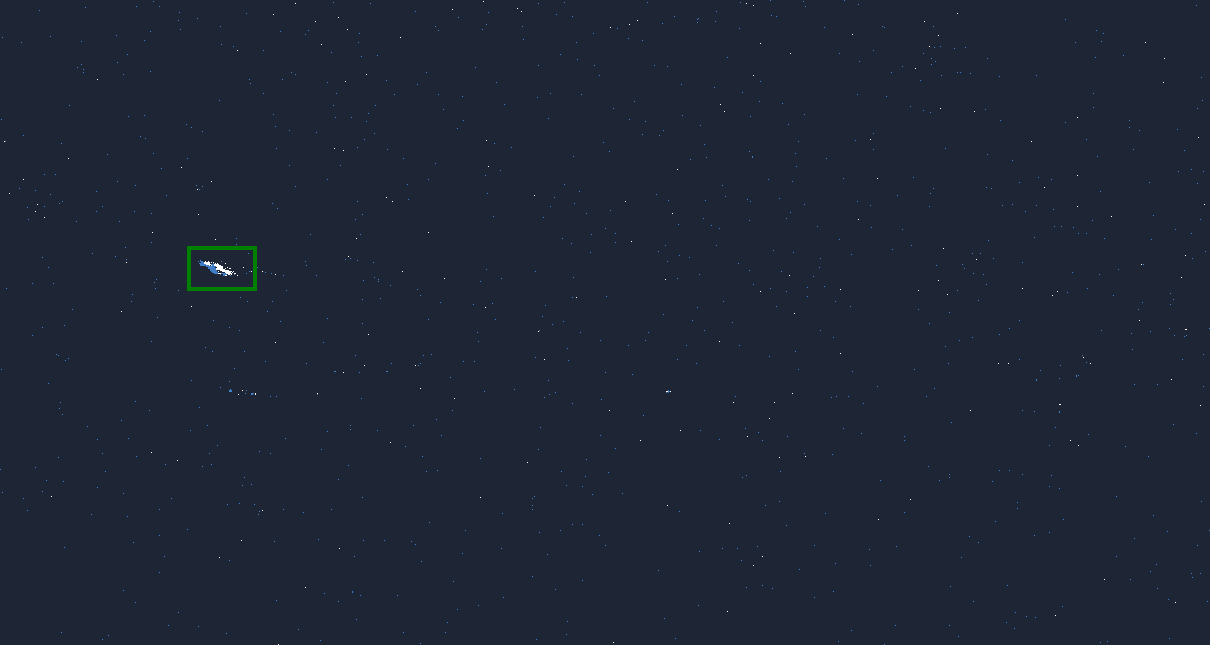} &
    \includegraphics[width=\figwidthnew,trim={150px 150px 150px 0},clip]{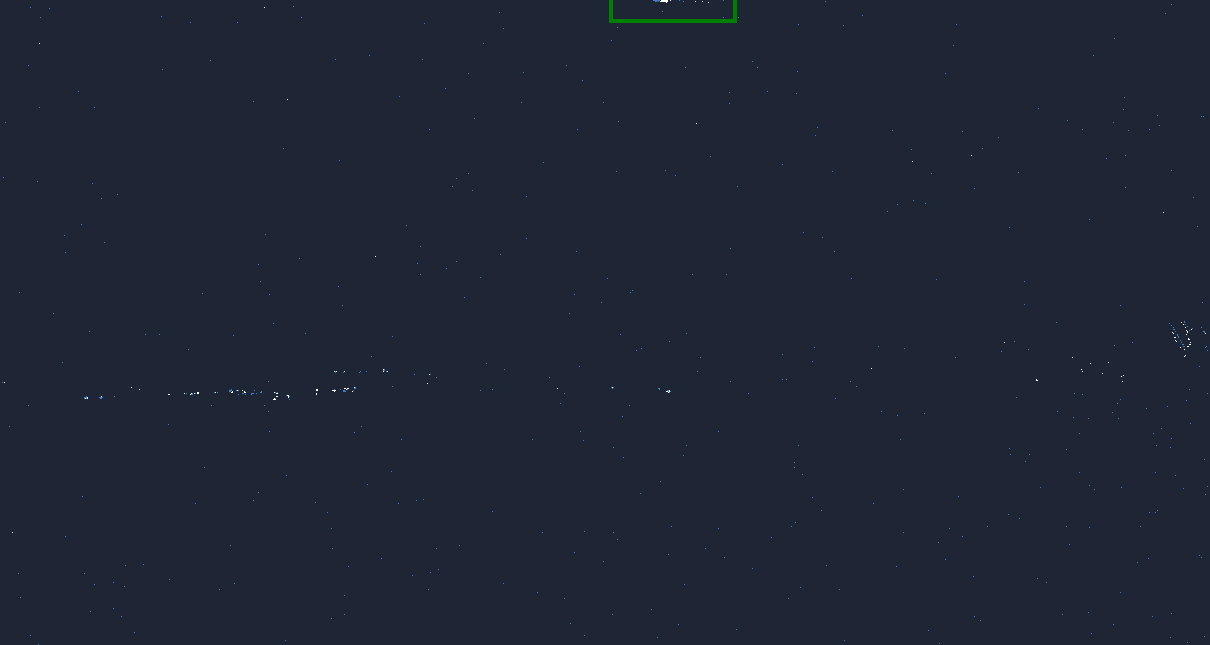} \\
    \end{tabular}

    \caption{Fixed recordings in an evolving scenario, highlighting the Event camera resilience to RGB domain shifts.}
    \label{fig:day2night_frames}
\end{figure*}

\newcommand{\figwidthc}{.19\textwidth}
\begin{figure*}[t]
    \centering
    \setlength{\tabcolsep}{1pt}
    \begin{tabular}{ccccc}
    Night & Shadows & Lens Flare & Insects & Rain \\
    \includegraphics[width=\figwidthc,trim={600px 200px 300px 250px},clip]{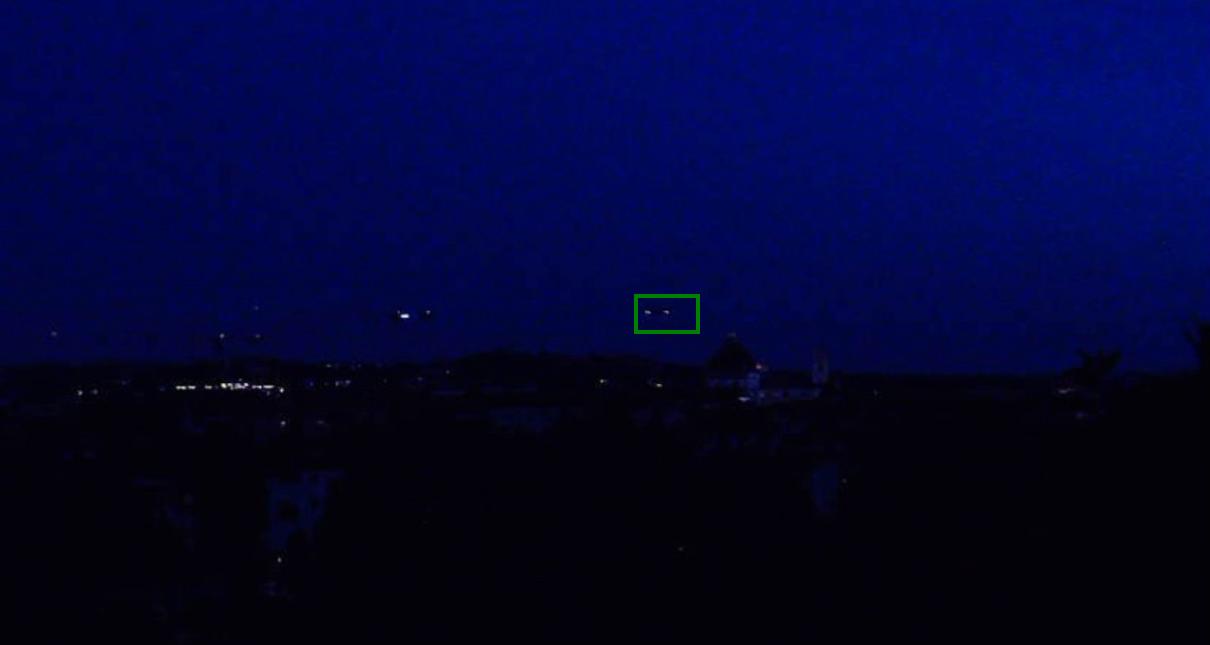} &
    \includegraphics[width=\figwidthc,trim={350px 350px 550px 100px},clip]{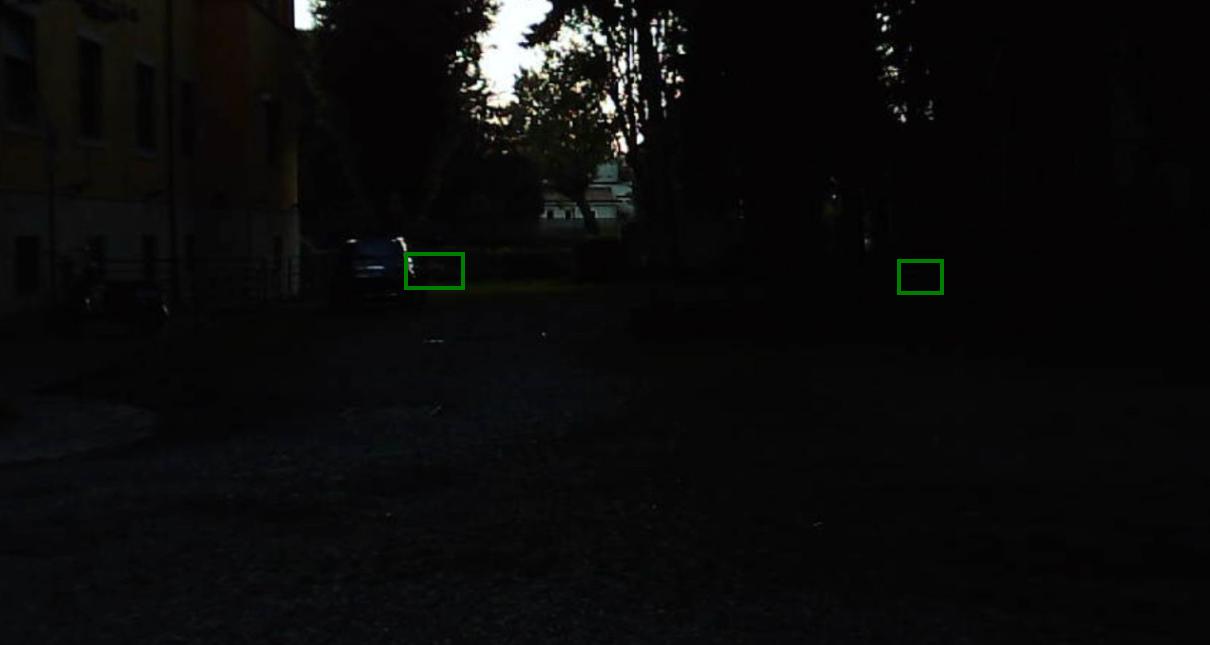} &
    \includegraphics[width=\figwidthc,trim={100px 300px 800px 150px},clip]{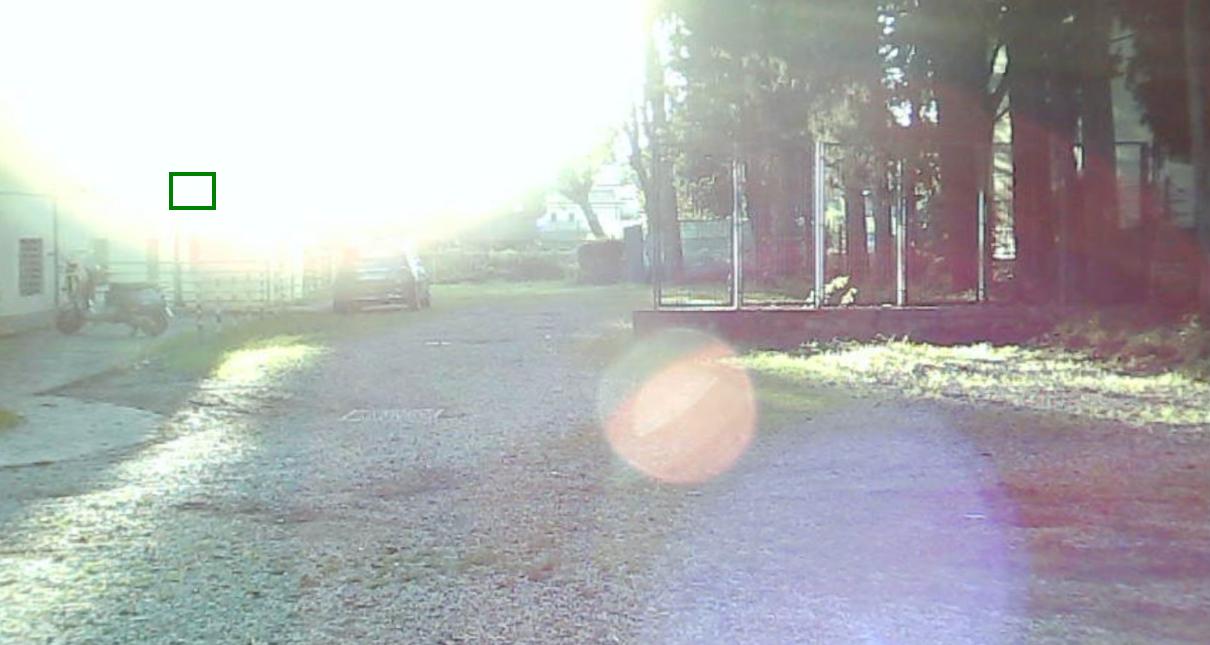} &
    \includegraphics[width=\figwidthc,trim={650px 300px 250px 150px},clip]{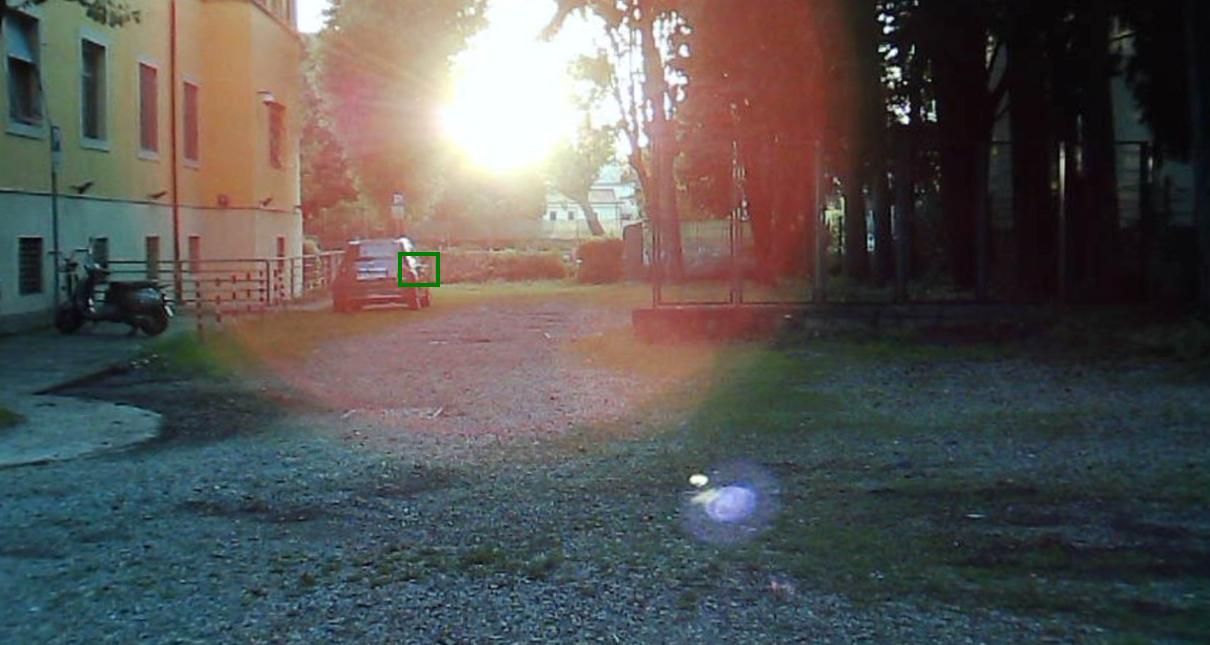} &
    \includegraphics[width=\figwidthc,trim={250px 450px 650px 0},clip]{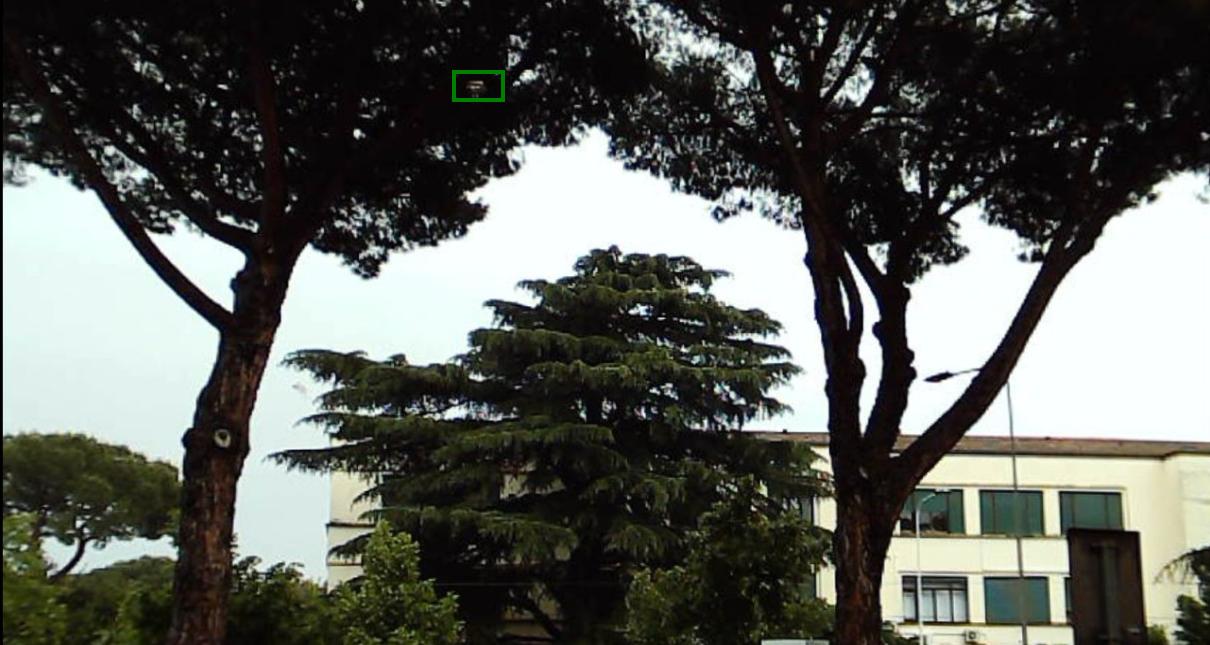} \\
    
    \includegraphics[width=\figwidthc,trim={600px 200px 300px 250px},clip]{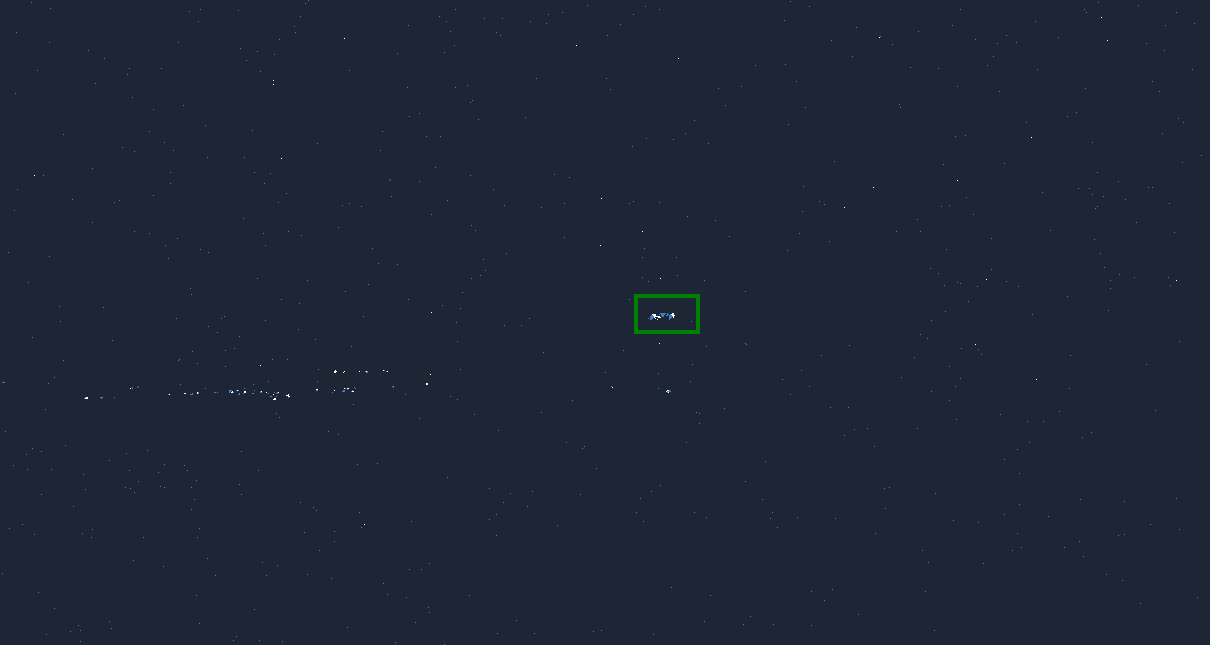} &
    \includegraphics[width=\figwidthc,trim={350px 350px 550px 100px},clip]{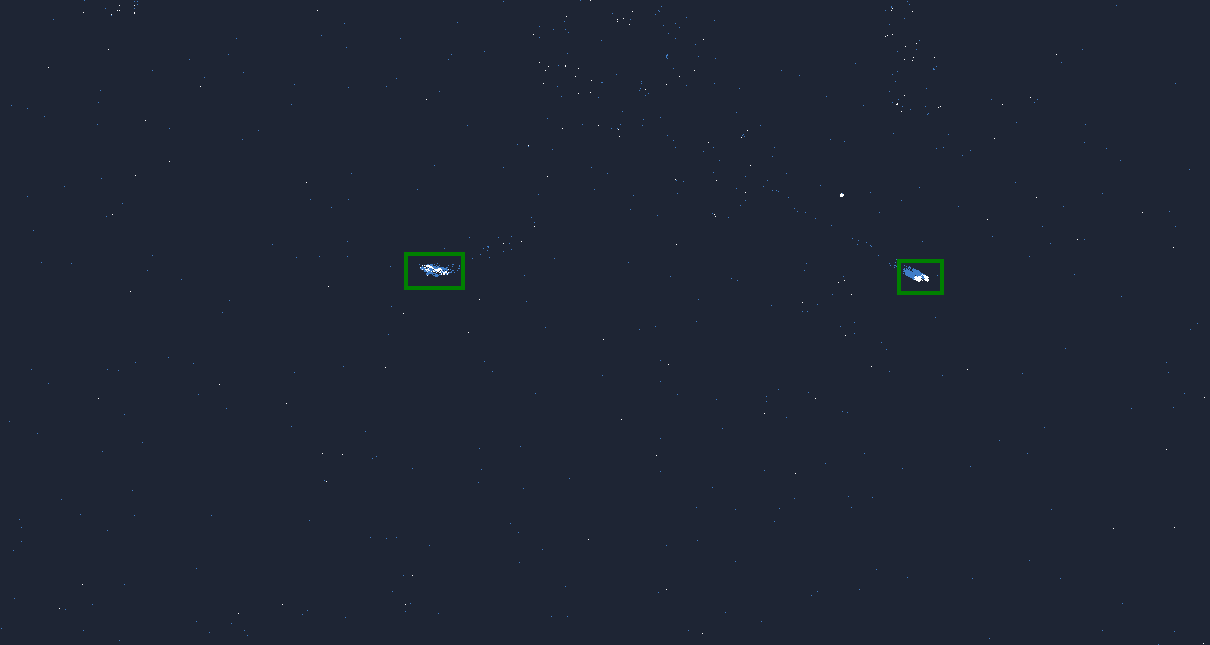} &
    \includegraphics[width=\figwidthc,trim={100px 300px 800px 150px},clip]{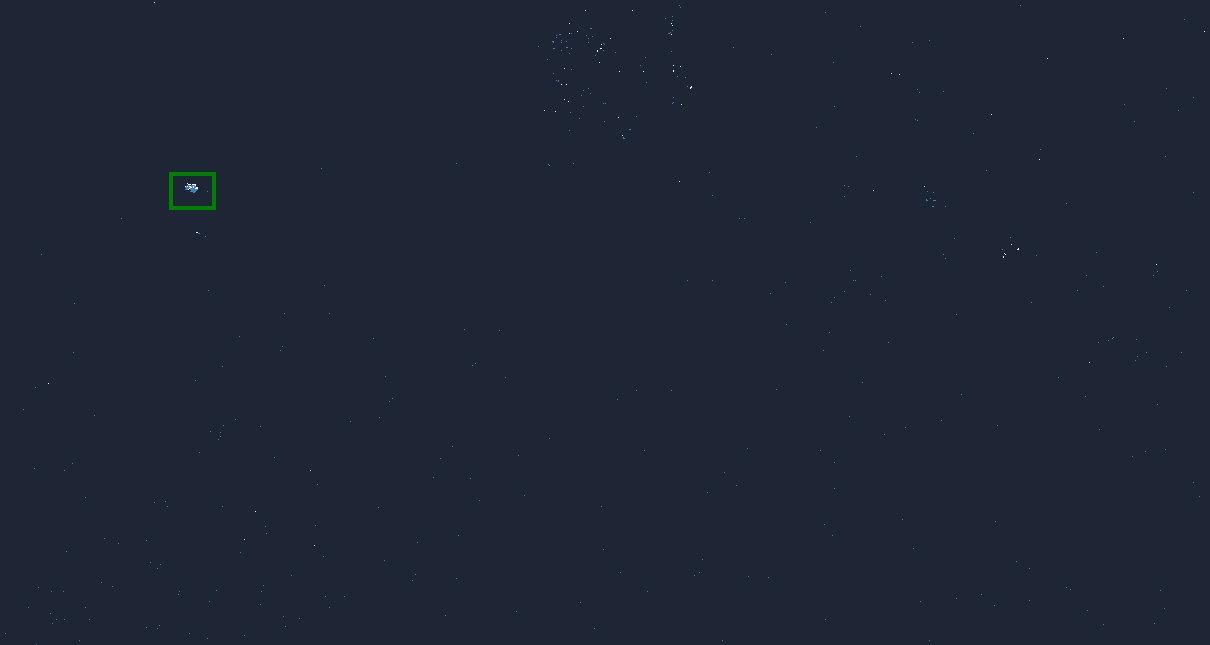} &
    \includegraphics[width=\figwidthc,trim={650px 300px 250px 150px},clip]{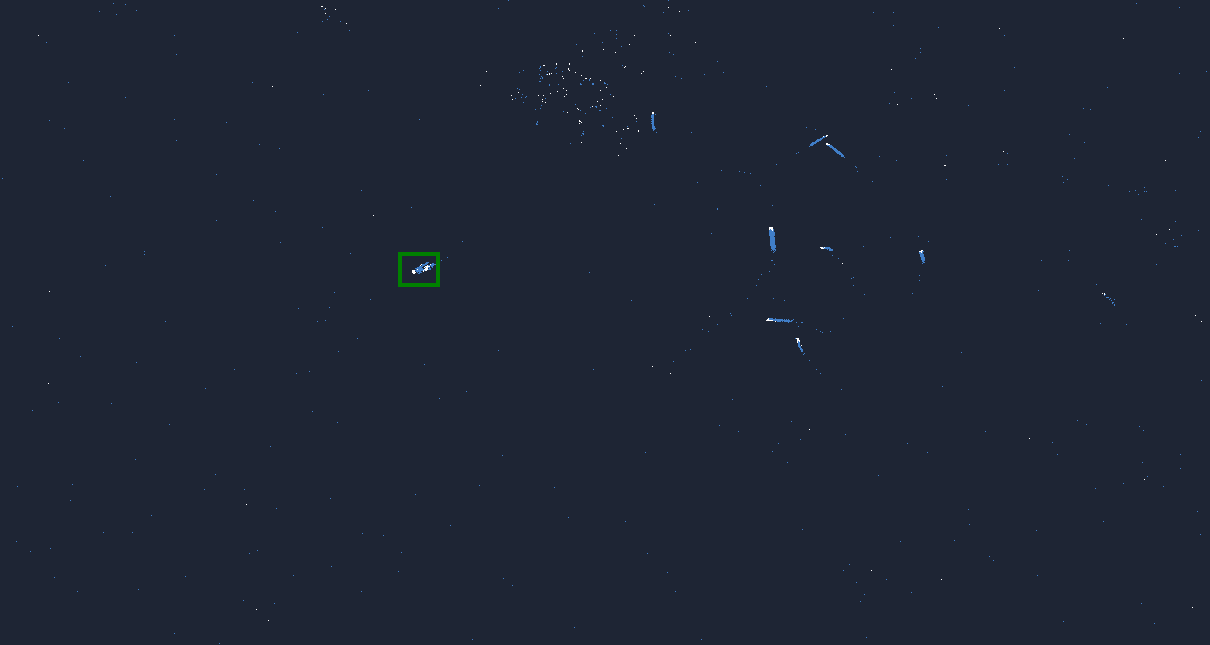} &
    \includegraphics[width=\figwidthc,trim={250px 450px 650px 0},clip]{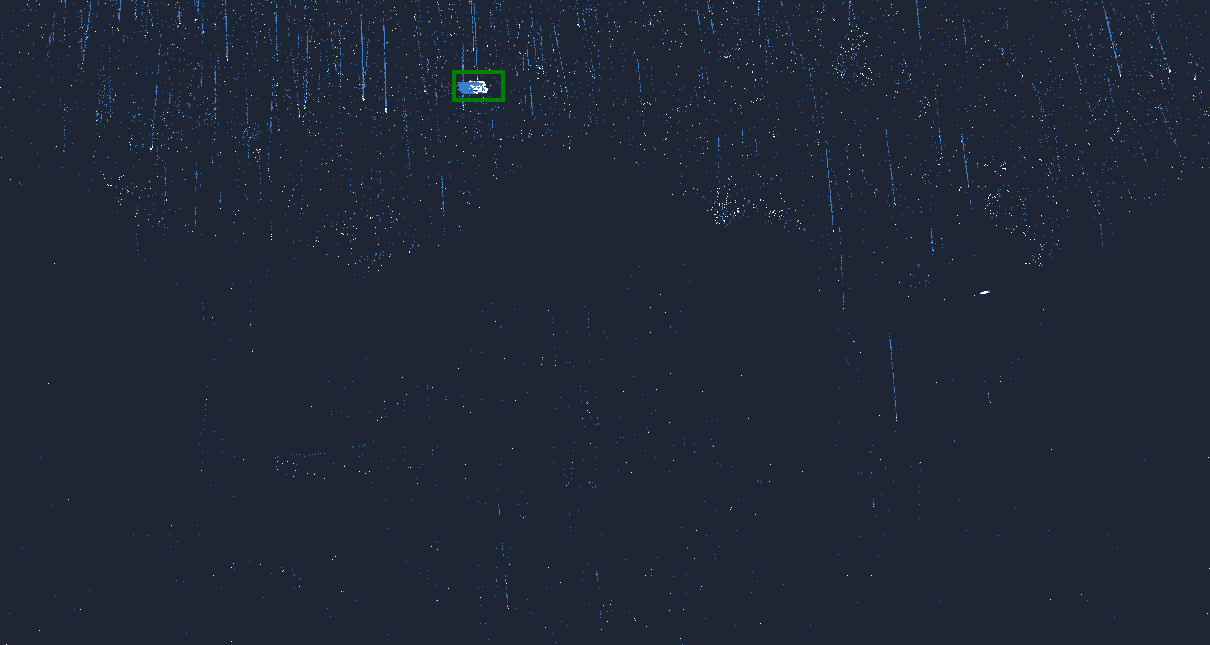} \\
    \end{tabular}

    \caption{Challenging frames for the RGB and/or event modality. With adverse lighting conditions (night, shadows, lens flare), the drone is not well visible in the RGB domain. Insects and rain act as distractors in the event domain.}
    \label{fig:hard_frames}
\end{figure*}

\section{The Florence RGB-Event Drone Dataset}
In the following, we present our dataset \dataset{}, discussing data collection and annotation and we present the related benchmarks.
%\subsection{Hardware}
\paragraph{\textbf{Sensors}}
To collect the dataset, we used a dual-camera single-mount spatially synchronized setup consisting of an HD RGB camera and an equivalent resolution Event camera. 
In particular, we opted for the Svpro HD camera with a varifocal lens for the RGB camera, presenting a resolution up to 1920x1280. For the Event-based camera we chose the Prophesee EVK4 HD mounted with an 8mm optic and using the Sony IMX636ES HD sensor. 
The mount consists of the two cameras on a tripod placed as close as possible to each other. Given the relatively small shift along the x-axis between the two cameras, particularly with respect with the target distance in the recorded scenes, we can safely assume that the resulting streams from the two cameras are sufficiently overlappable given some minor cares. 
In fact, to ensure a better overlapping, we first computed the intrinsic parameters of both cameras, removing the distortion. Then, we added a minor padding around the RGB camera image to better match the x-axis shift. Coupled together, these result in two domains with a shared reference system, meaning that a given object in the scene will have the same pixel-wise coordinates in both recordings. 

To temporally synchronize the two modalities, we opted for a mixture of software and handmade corrections. 
In particular, the RGB camera has been fixed to 30 FPS for all recordings, where each video lasts approximately 120 seconds.
The event camera, on the other hand, is not constrained by frames per second; in fact, it provides arbitrarily precise frame creation. This means that we can extract frames at the same recording frequency of the RGB counterpart, and achieve temporal synchronization once we align the RGB and Event camera recordings starting point. We release both the original aligned event file format and the extracted frames at the same RGB camera frequency, for an easier usage of the dataset. 
The result is a dataset with corresponding start and end for all videos, making them temporally overlappable as well.

\paragraph{\textbf{Drones}}
We recorded 5 different models of drones, ranging from mini-drones to commercially available models. 
Two of these drones are FPV-controlled and weigh under 50 grams each, maneuvered using a Radiomaster Pocket with Elrs protocol as remote controller and the Eachine ev300d analogic FPV goggles, enabling precise movements and non-trivial trajectories. Both of them are also GPS equipped, for additional stabilization. The smallest one is a Betafpv air75 measuring 115x115x52mm, with a weight of 30g and a maximum recorded speed of 100km/h.
The other mini-drone is a DarwinFPV cineape20 measuring 125x125x90mm, weighing 100g and with the maximum speed recorded of 50km/h.  
These drones, given their precision and maneuverability, can mimic the trajectories of biological organisms such as flying insects, leading to significant challenges, particularly in the event-based camera domain.
A significant characteristic of such mini-drones is the difficulty of maintaining stable hovering, for which we deemed it necessary to include other drones to also cover drone hovering scenarios.
A medium-sized commercial drone has also been used, in particular a DJI Tello EDU quadcopter, weighing 87g and with dimensions of 98×92.5×41mm. 
All of these small-to-mid-sized drones are also particularly interesting for their susceptibility to atmospheric phenomena (e.g., wind), while also posing non-trivial challenges regarding their detection even at moderate distances. 
Finally, two bigger commercial drones have also been recorded, namely the DJI Mini 2 and DJI Mini 3. 
Both of these drones weigh 249g and can reach up to 57.6 km/h, with the Mini 2 having dimensions of 245×290×55mm, while the Mini 3 reaches 251×362×72 mm. 
These two models give the possibility of more stable hovering and much more distant recordings, reaching more than 60 meters from the observer.

\paragraph{\textbf{Annotation}}
To annotate the dataset, we leveraged the sparse nature of the event frames to facilitate the task. The annotations are made on event frames, obtained using Prophesee's built-in frame extractor at 30 FPS to match the frame-rate of the RGB stream. The resulting annotations are also valid for the RGB videos, thanks to the spatio-temporal synchronization.
We built a framework consisting of 3 modules: a preliminary automatic annotator based on an algorithmic spatter tracking~\footnote{\url{https://docs.prophesee.ai/stable/samples/modules/analytics/tracking_spatter_py.html}}, a custom-made software correcting annotations, and an interpolation module to ensure track continuity and refine box details using the temporal context. A final manual inspection is carried out to check the quality of the annotations.
%\textcolor{red}{To further expand the dataset tasks and quality, we also manually re-annotated all the integrated NeRDD dataset double-drones videos, adding a new "id" column, representing the drone identity in the video, to extend it for other tasks and scenarios.}
All the annotated drones have a unique ID within each video. We leverage these annotations to go beyond simple drone detection, enabling the development and evaluation of tracking and forecasting models.
%To make possible both the annotation phase and the correct box projection to the RGB domain, we first need to extract the event based frames. We used Prophesee's built-in frame extractor to generate 30 fps frames, or the same as the RGB domain.
%Once we obtained the initial boxes from the spatter tracker, we proceeded to manually fix frame-by-frame all the missing or excessive annotations, using the custom made annotation software. At the end of this stage, we once again performed a manual quality check on all recordings. 
%Finally, the interpolation modules ensure continuity and better boxes by interpolating the manually annotated and automatically extracted bounding boxes. This creates consistent and continuous boxes. 

\paragraph{\textbf{The \dataset{} Dataset}}
As a result of the recording and annotation phase, we achieved a total of more than 7 hours per modality (14+ hours combined) of annotated drone recordings in different environments and scenarios.
%The dataset is comprised of a total of 7 hours of bimodal recordings; this means that considering both modalities we reach 14 hours of combined videos. 
%In particular, we summarize the dataset qualities in \todo{do tab}. 
Among these, we identify sub-categories of videos recorded in precise scenarios: of the total recording time, 40 minutes contain multiple drone flights in various locations and scenarios; rain is present in 20 minutes of recordings, while low-light scenarios account for more than 1 hour, and indoor recordings cover around 24 minutes. These scenarios are not mutually exclusive. %, as the dataset presents extremely challenging situations in which both RGB and event-based are hindered by adverse conditions.

While recording the dataset, we focused on creating as many varying scenarios as possible, not only to help models generalize better on the tasks proposed, but also to better investigate and research the complementary properties of event-based and RGB domains.
To do so, we decided to present two separate splits for our dataset, a canonical one and a challenging one.

The \textit{canonical split} is obtained by simply partitioning the data into train and validation with an 80/20 split. We carefully ensured to balance between different scenarios in the two splits: scenarios like nighttime, rainy background or with two simultaneous flying drones are present equally in both train and validation. 
This split makes a good benchmark for generic models in both domains. 

The \textit{challenging split} contains the same scenarios in both training and validation, but presents a clear shift in data distribution of either one or both domains. While the scene is still present in both training and validation, variations in the environment create a significant shift in how the scenario is perceived in either domain. For example, a steep decrease in brightness on the scene, as around sunset, greatly impacts the RGB camera output (see Fig.~\ref{fig:day2night_frames}); conversely, rain or the presence of small flying insects may significantly change the event camera output, even if the scenario and background are unchanged. These cases can bring serious challenges for deep learning methods, while at the same time representing a more realistic indicator of a model's generalization capabilities. Some examples of challenging scenarios are shown in Fig.~\ref{fig:hard_frames}. Video samples can be found on the dataset webpage: \url{https://miccunifi.github.io/FRED/}.

\section{Benchmarks}

In the following, we provide a formal definition of the proposed benchmarks. We first define the inputs, and we then describe each task individually.
% --------------------- Definition of event stream --------------------- 
Let $\mathcal{E}$ denote a stream of events $\mathcal{E} = \{e_i = (x_i, y_i, p_i, t_i)\}_{i \in \mathbb{N}}$
where $x_i \in [0, W-1]$, $y_i \in [0, H-1]$ are the spatial coordinates of the event $e_i$, $p_i \in \{0, 1\}$ is its polarity and $t_i \in [0, \infty)$ its timestamp.
% --------------------- Definition of video stream --------------------- 
Let $\mathcal{V}$ denote a video stream of synchronous frames $\mathcal{V} = \{(f_j, t_j)\}_{j \in \mathbb{N}}$ captured at a frame rate of $\frac{1}{T}$ FPS, where each $f_j$ is the $j$-th RGB frame of size $(W, H, 3)$ and $t_j=jT$ is its corresponding timestamp.

\textbf{\textit{Drone Detection.}}
% --------------------- Drone detector --------------------- 
The goal of the drone detection task is to train a drone detector model $\mathcal{D}(\mathcal{I})$ that
% --------------------- detector output --------------------- 
outputs a set of bounding boxes $\hat{B} = \{\hat{b}_k = (\hat{x}_k, \hat{y}_k, \hat{w}_k, \hat{h}_k, \hat{t}_k)_{k \in \mathbb{N}} \}$, where each box is identified by its top-left coordinate $(\hat{x}_k, \hat{y}_k)$, its width $\hat{w}_k$ and height $\hat{h}_k$ and is temporally localized by its timestamp $\hat{t}_k$.
% --------------------- detector input --------------------- 
The detector processes an input $\mathcal{I}_t \in [\mathcal{E}|\mathcal{V}]_{t-\Delta:t}$ composed of event or RGB data (or both) acquired in a time-frame over a temporal interval $\Delta$ before the detection time $t$. In other words, a detection at time $t$ must rely solely on data accessible before $t$, without accessing future information.
% --------------------- detector evaluation --------------------- 
Following the evaluation protocol of \cite{perot2020learning}, we assess the detector at fixed time intervals of size $\tau$. In general, $\Delta$ and $\tau$ may differ; in particular, $\Delta > \tau$ implies that inputs for adjacent detection times may overlap. For simplicity, we set both parameters to the inverse of the RGB camera's framerate, i.e., $\Delta = \tau = 33$ ms.

% --------------------- metrics --------------------- 
To evaluate performance, we employ standard object detection metrics, namely mean Average Precision (mAP) at Intersection over Union (IoU) threshold 0.5 (mAP$_{50}$), and averaged across thresholds from 0.5 to 0.95 with a step of 0.05 (mAP$_{50:95}$)~\cite{lin2014microsoft}.

\textbf{\textit{Drone Tracking}.}
% --------------------- Drone Tracker --------------------- 
The goal of the drone tracking task is to train a tracker model $\mathcal{T}$ that, at each discrete timestep $t \in \{0, \tau, 2\tau, \dots\}$, maintains a set of temporally consistent drone trajectories.
The tracker receives as input all sensor data acquired in the interval $(t - \tau, t]$, denoted as $\mathcal{I}_t \in [\mathcal{E}|\mathcal{V}]_{t-\tau:t}$. Depending on the relative sampling frequencies of the sensors and $\tau$, one or both modalities may be available at each step.
For example, if RGB frames are captured less frequently than the tracking interval $\tau$, no RGB data may be present in certain intervals, while event data, being asynchronous and dense, is assumed to be continuously available.

% --------------------- Tracker output  --------------------- 
At each timestep $t$, the tracker must produce an estimate of the set of active object tracks
$\hat{S}_t = \{\hat{s}_m = (\hat{x}_m, \hat{y}_m, \hat{w}_m, \hat{h}_m, \hat{id}_m)\}_{m \in \mathbb{N}_t}$,
where each track $\hat{s}_m$ consists of a bounding box $(\hat{x}_m, \hat{y}_m, \hat{w}_m, \hat{h}_m)$ and a unique identity label $\text{id}_m$.
In other words, the tracker is expected to update the set of active tracks $\hat{S}_t$ by either a) refining the state of existing tracks; b) initializing new tracks when new drones appear; c) terminating tracks that are no longer observable.
The tracker must operate causally, relying only on past and present data (i.e., data in $(t - \tau, t]$) and without access to future sensor observations.

% --------------------- Tracking eval  --------------------- 
Tracking is evaluated at each discrete timestep. Similarly to the detection task, we set $\tau = 33ms$ to match the RGB camera framerate.
At each timestep, tracker outputs are compared to ground-truth annotations using standard multi-object tracking metrics such as MOTA (Multi-Object Tracking Accuracy), IDF1 (Identity F1 score), ID Switch, Precision and Recall~\cite{bernardin2008evaluating}.
Ground-truth annotations consist of per-frame bounding boxes and consistent object identities for all visible drones, labeled at the same frequency as the evaluation interval. Drones that exit the field of view and re-enter are considered as separate tracks.

\begin{table}[t]
    \centering
    \resizebox{0.65\columnwidth}{!}{\begin{tabular}{l|c c|c c|c c c c}
         & & & \multicolumn{2}{|c|}{Canonical Split} & \multicolumn{4}{c}{Challenging Split} \\ \hline 
        Method & Event & RGB & mAP$_{50}$ & mAP$_{50:95}$ & mAP$_{50}$ & mAP$_{50:95}$ \\ \hline
        YOLO \cite{khanam2024yolov11} & \checkmark
 & $\times$ & \textbf{87.68} & \textbf{49.25} & \textbf{79.60} & \textbf{41.63} \\
        YOLO \cite{khanam2024yolov11} & $\times$ & \checkmark
 & 35.24 & 13.40 & 18.13 & 6.33 \\
        RT-DETR \cite{zhao2024detrs} & \checkmark
 & $\times$ & 82.05 & 38.98 & 76.93 & 35.08 \\
        RT-DETR \cite{zhao2024detrs} & $\times$ & \checkmark
 & 34.12 & 11.79 & 21.12 & 7.13 \\
 Faster-RCNN \cite{ren2015faster} & \checkmark
 & $\times$
 & 85.00 & 43.40 & 75.40  & 34.90 \\
 Faster-RCNN \cite{ren2015faster} & $\times$
 & \checkmark
 & 35.10 & 12.30 & 16.40 & 4.80 \\
 ER-DETR \cite{magrini2024neuromorphic} & \checkmark
 & $\times$
 & 68.42 & 21.80 & 62.20 & 19.81 \\
 ER-DETR \cite{magrini2024neuromorphic} & $\times$
 & \checkmark
 & 28.56 & 7.73 & 15.84 & 4.11 \\
 ER-DETR \cite{magrini2024neuromorphic} & \checkmark
 & \checkmark
 & 78.59 & 32.21 & 75.73 & 27.87\\
    \end{tabular}}
    \caption{Detection results.}
    \label{tab:detection}
\end{table}

\begin{table}[t] % Use table* for full page width if needed, or table for column width
    \centering
   \resizebox{0.65\columnwidth}{!}{ 
    \begin{tabular}{l|cc|ccHcccHHHHHH}
        %& & & \multicolumn{6}{|c|}{Canonical Split} & \multicolumn{6}{c}{Challenging Split} \\ \hline 
        Method & Event & RGB & MOTA $\uparrow$ & IDF1 $\uparrow$ & MOTP $\uparrow$ & IDs $\downarrow$ & P $\uparrow$ & R $\uparrow$ & MOTA $\uparrow$ & IDF1 $\uparrow$ & MOTP $\uparrow$ & IDs $\downarrow$ & P $\uparrow$ & R $\uparrow$ \\ \hline
        YOLO \cite{khanam2024yolov11} & \checkmark & $\times$  & 57.1 & \textbf{47.6} & 29.5 & 436 & 83.0 & 72.2 & 39.3 & 39.9 & 32.0 & 832 & 73.7 & 61.9  \\ 
        YOLO \cite{khanam2024yolov11} & $\times$ & \checkmark & 21.5 & 28.7 & 31.4 & 9830 & 75.7 & 32.2 &  &  &  &  &  &     \\
        RT-DETR \cite{zhao2024detrs} & \checkmark & $\times$ & 51.3 & 45.1 & 31.2 & \textbf{301} & 77.4 & 72.8 &  &  &  &  &  &    \\
        RT-DETR \cite{zhao2024detrs} & $\times$ & \checkmark & 10.1 & 30.0 & 32.4 & 692 & 57.1 & 42.6 &  &  &  &  &  &    \\
        Faster-RCNN \cite{ren2015faster} & \checkmark & $\times$ & \textbf{64.0} & 45.0 & 25.6 & 4191 & \textbf{83.8} & \textbf{83.1} &  &  &  &  &  &    \\
        Faster-RCNN \cite{ren2015faster} & $\times$ & \checkmark & -23.1 & 24.7 & 31.0 & 2298 & 40.7 & 46.8 &  &  &  &  &  &    \\
        ER-DETR \cite{magrini2024neuromorphic}  & \checkmark & $\times$ & 45.1 & 42.1 & 32.9 & 476 & 74.3 & 69.5 &  &  &  &  &  &    \\
        ER-DETR \cite{magrini2024neuromorphic} & $\times$ & \checkmark &  -13.4 & 0.5 & \textbf{34.3} & 2345 & 13.3 & 2.1 &  &  &  &  &  &    \\
        ER-DETR \cite{magrini2024neuromorphic} & \checkmark & \checkmark & 46.2 & 43.3 & 32.7 & 431 & 75.5 & 69.0 &  &  &  &  &  &    \\
    \end{tabular}
    }
    \caption{Multiple Object Tracking results.}
    \label{tab:mot_performance}
\end{table}

\begin{table}[t]
    \centering
    \resizebox{\columnwidth}{!}{ 
    \begin{tabular}{l|cc|ccc|ccc}
    \multicolumn{3}{c}{} & \multicolumn{3}{c}{Short-term ($H_p=0.4s$)} & 
    \multicolumn{3}{c}{Mid-term ($H_p=0.8s$)} \\ \hline
    Method & Event & RGB & (ADE/FDE)$_{BB} \downarrow$ & (ADE/FDE)$_{C} \downarrow$ & mIoU $\uparrow$ & (ADE/FDE)$_{BB} \downarrow$ & (ADE/FDE)$_{C} \downarrow$ & mIoU $\uparrow$  \\
    \hline
    %Kalman Filter & $\times$ & $\times$ & ? & ? & ? & ? & ?\\
    LSTM & $\times$ & $\times$ &82.73/90.34  & 171.9/53.04 & 0.044 &105.6/145.9 & 353.3/96.44 & 0.027 \\
    Transformer & $\times$ & $\times$ & 49.43/69.07 & 124.8/46.74 & 0.283 & 76.23/125.9 & 292.2/87.68 & 0.193\\
    CNN+Transformer & \checkmark & $\times$ & 48.46/66.67 & 124.0/45.92 & \textbf{0.284}  & 72.85/\textbf{119.6} & \textbf{280.9}/\textbf{83.86} & \textbf{0.206} \\
    %TODO Drone & Forecasting & ? \\
    CNN+Transformer & $\times$ & \checkmark & 59.39/83.86 & 151.7/58.06 & 0.191 & 88.46/143.5 & 337.0/100.6 & 0.136\\
    CNN+Transformer & \checkmark & \checkmark & \textbf{47.49}/\textbf{65.89} & \textbf{121.3}/\textbf{45.23} & 0.279 & \textbf{72.81}/122.5 & 281.5/85.78 & 0.201 \\
    \end{tabular}
    }
    \caption{Trajectory forecasting results. ADE and FDE results are in pixels.}
    \label{tab:forecasting_res}
\end{table}  

\textbf{\textit{Drone Trajectory Forecasting}.}
\label{subsub:forecasting}
The goal of the drone forecasting task is to predict the future motion of visible drones based solely on past information available up to a given reference time $t$. Each sample is anchored at a discrete timestep $t$, referred to as the \emph{forecasting time}, which defines the boundary between observable history and the prediction horizon.
%The forecasting model can only access data from times $t' \leq t$, and must generate predictions for times $t' > t$.

Let $b_m^{(t')} = (x_m^{(t')}, y_m^{(t')}, w_m^{(t')}, h_m^{(t')})$ denote the bounding box of drone $m$ at time $t'$, represented by its top-left coordinate $(x, y)$, width $w$, and height $h$. For each drone $m$ visible at time $t$, we define its past trajectory as a sequence of bounding boxes over the observation horizon $H_p=\tau N_p$, discretized into $N_p$ timesteps of size $\tau$:
%\begin{equation}
$B_m^{\text{p}} = \left\{ b_m^{(t - N_p \tau + i \tau)} \right\}_{i = 0}^{N_p - 1}$.
%\end{equation}
The forecasting task requires predicting the future trajectory over a forecast horizon $H_f = \tau N_f$ of $N_f$ steps with interval $\tau$:
%\begin{equation}
$\hat{B}_m^{\text{f}} = \left\{ \hat{b}_m^{(t + i \tau)} \right\}_{i = 1}^{N_f}$.
%\end{equation}
% The model may optionally be conditioned on sensor data observed within the same past window $[t - (N_p - 1)\tau, t]$, including:
% \begin{itemize}
%     \item RGB frames from the video stream $\mathcal{V}$;
%     \item Event data from the asynchronous stream $\mathcal{E}$.
% \end{itemize}
Each sample may include input data observed in the past interval $[t-H_p, t]$ from either the event modality $\mathcal{E}$ or the RGB modality $\mathcal{V}$ (or both).
%Note that due to differing frame rates and modalities, not all sensor inputs may be available at every timestep.

We evaluate forecasting quality using Average Displacement Error (ADE) and Final Displacement Error (FDE), i.e. the L2 distance between predictions and ground truth respectively for the whole sequence and the last prediction, computed over the forecasted bounding boxes and their centers.
In addition, to evaluate the geometric overlap between predicted and ground-truth boxes, we compute the Intersection over Union (IoU) at each future timestep and average across the horizon (mean IoU - mIoU).
For the forecasting benchmark, we use an observation horizon $H_p=0.4s$ and two different prediction horizons: $H_f=0.4s$ for short-time predictions and $H_f=0.8s$ for mid-term predictions.
For this task, we use the canonical split of \dataset{} by sampling almost 5K different 1.2s-long trajectories (3680 for train and 940 for test).

\section{Experiments}

\textbf{\textit{Drone Detection}.}
Tab. \ref{tab:detection} presents the detection results obtained on the canonical and challenging splits of \dataset{} with four state-of-the-art detectors: YOLO v11 \cite{khanam2024yolov11}, RT-DETR \cite{zhao2024detrs}, Faster-RCNN~\cite{ren2015faster}, and the neuromorphic Event-RGB DETR (ER-DETR) \cite{magrini2024neuromorphic}.
For every model, we report mAP$_{50}$ and  mAP$_{50:95}$ obtained with event frames, RGB frames, and, for ER-DETR only, their fusion.

Across both splits, models trained on the event stream achieve markedly higher accuracy than their RGB counterparts, confirming the importance of such modality.
All single‑modality detectors experience a drop in performance on the challenging split. Low-light sequences, airborne distractors such as insects and different weather conditions represent as a distribution shift, making the task harder for the detectors.
Fusing the two modalities alleviates part of this degradation. ER‑DETR with joint Event+RGB input maintains almost the same performance across the two splits. This suggests that the two modalities are complementary: RGB supplies texture when the drone is nearly stationary and event activity is sparse, while events preserve shape under adverse illumination and cluttered backgrounds. Nevertheless, the fusion model does not yet surpass the strongest event‑only baseline (YOLO on events) on the canonical split, indicating that current fusion strategies leave room for further optimization.

%By highlighting modality‑specific failure modes and the limits of existing fusion techniques, \dataset{}  establishes a focused benchmark for resilient aerial perception. The performance gaps observed here motivate research on domain adaptation, cross‑modal attention and other methods that can more effectively exploit complementary and synergetic information from two modalities.

\textbf{\textit{Tracking}.}
We leverage the outputs of the object detectors to perform tracking-by-detection with the state-of-the-art tracker ByteTrack~\cite{zhang2022bytetrack}. We report the results in Tab.~\ref{tab:mot_performance}, obtained by performing tracking on the whole test set of the canonical split.
Although there is no clear winner against the competing methods, a few interesting insights emerge clearly. First, the results confirm that tracking drones relying only on the RGB domain is an extremely challenging task. Second, the multimodal version of ER-DETR slightly improves upon its event-only counterpart.
The overall best-performing method is Faster-RCNN, although it exhibits an extremely high number of ID switches. These results act as a starting baseline for future works. In particular, multimodal approaches combined with strong modern architectures like YOLO appear to be an interesting direction to pursue.

\textbf{\textit{Forecasting}.}
In Tab. \ref{tab:forecasting_res} we report the results on the forecasting task. We compare different baselines on the canonical split of \dataset{}, including \textit{blind} ones that have no visual input but rely on the box information alone.
%In all our forecasting experiments, we set a 0.4s time interval for the input time (past trajectory) and 0.8s of prediction (future trajectory).
All models are trained to minimize $\mathcal{L} = L_{N_f}+\lambda L_{N_p}$, with
%\begin{equation}
$L_{N_x} = \frac{1}{N_x} \sum_{i=1}^{N_x} \sqrt{ \left( \hat{b}_m^{(t + i\tau)} - b_m^{(t + i\tau)} \right)^2}$
%\end{equation}
and $\lambda$ is a scaling coefficient used to weigh the input reconstruction term (in our experiments $\lambda=0.5$).

%\paragraph{Kalman Filter} ...
%\textit{LSTM.} 
The first baseline consists of a Long Short-Term Memory (LSTM) network~\cite{hochreiter1997long} that takes as input only the bounding box coordinates of the tracked object at each timestep.
%\textit{Transformer.}
Similarly, we train a Transformer model operating solely on bounding box coordinates of the tracked object at each timestep. The encoder processes the sequence of past bounding box locations, and the decoder predicts both the past (reconstruction) and the future trajectory (forecasting). Both the encoder and the decoder have 4 heads and 4 layers each.
%\textit{Transformer+CNN.}

To incorporate visual information, we extended the Transformer architecture by integrating features extracted from a simple 3-layer Convolutional Neural Network (CNN). In this model, for each timestep, we crop the image based on the bounding box coordinates and feed this crop to the CNN. The resulting feature vector is then concatenated with the bounding box features before being fed into the encoder-decoder Transformer. We train variants with an RGB encoder, an event encoder, and both.

The results of Tab. \ref{tab:forecasting_res} are presented in terms of ADE and FDE in pixels for the original image size of $1280\times720$.
As expected, the baselines with no visual input (LSTM and Transformer) perform worse than the rest. The use of the event modality (either alone or alongside RGB) shows an increase in all the metrics for both tested time horizons.

\section{Conclusions}
We presented the \dataset{} dataset for benchmarking neurmorphic and RGB drone perception.
It is designed for drone detection, tracking, and trajectory forecasting, combining RGB video and event streams, which are spatio-temporally synchronized. The dataset comprises over 7 hours of high-resolution (1280x720 HD) recordings per modality, making it the largest existing drone detection dataset, as well as the first one for drone tracking and forecasting.
A key strength of FRED lies in its diverse and challenging scenarios, including adverse weather, distractors such as insects, and varying lighting conditions such as day-to-night transitions. It features recordings of five different drone models, ranging from mini-drones to commercial models, exhibiting diverse maneuverability and speed capabilities.
In the paper, we provided baselines for each proposed task, hoping to foster advancements in the field.

% ---------------------  --------------------- 

\bibliographystyle{plain}
\bibliography{bib}
\end{document}